\ifdefined\LaTeXMLversion
  \documentclass[12pt]{report}
  \newcommand{\bcshon}{}
  \newcommand{\defenceday}[1]{}
  \newcommand{\defencemonth}[1]{}
  \newcommand{\defenceyear}[1]{}
  \newcommand{\convocation}[2]{}
  \newcommand{\supervisor}[1]{}
  \newcommand{\reader}[1]{}
  \newcommand{\frontmatter}{}
  \newcommand{\mainmatter}{}
  \newenvironment{acknowledgements}{\chapter*{Acknowledgements}}{}
\else
  \documentclass[12pt, final]{dalcsthesis}
\fi
\usepackage{graphicx}
\usepackage{subcaption}
\usepackage{amsmath}
\usepackage{algorithm}
\usepackage[noend]{algpseudocode}
\usepackage{hyperref}
\usepackage{booktabs}
\usepackage{multirow}
\usepackage{mathptmx}
\usepackage{courier}
\usepackage{array}
\usepackage{listings}
\usepackage{longtable}
\usepackage{tablefootnote}

\newcommand{\resultsheader}{\textbf{Generation} & \textbf{Max} & \textbf{Mean} & \textbf{Median} & \textbf{Min} \\\hline}
\ifdefined\LaTeXMLversion
  \newcommand{\resultstable}[3]{%
    \begin{table}[!htb]\centering\caption{#1}\label{#2}%
    \begin{tabular}{|l|l|l|l|l|}\hline\resultsheader#3\end{tabular}%
    \end{table}}
\else
  \newcommand{\resultstable}[3]{%
    \begin{longtable}{|l|l|l|l|l|}
    \caption{#1}\label{#2}\\\hline\resultsheader\endfirsthead
    \hline\resultsheader\endhead
    \hline\endfoot
    #3\end{longtable}}
\fi

\graphicspath{{assets/experiments/aggregate_results/figures/}{assets/experiments/baseline/figures/}}

\begin{document}
\bcshon
\title{Reinforced Linear Genetic Programming}
\author{Urmzd Mukhammadnaim}
\defenceday{12}
\defencemonth{April}
\defenceyear{2023}
\convocation{May}{2023}

\supervisor{M. Heywood}
\reader{N. Zincir-Heywood}

\frontmatter
\addcontentsline{toc}{chapter}{List of Algorithms}
\listofalgorithms
\lstlistoflistings

\begin{abstract}
	Linear Genetic Programming (LGP) is a powerful technique that allows for a variety of problems to be solved using a linear representation of programs. However, there still exists some limitations to the technique, such as the need for humans to explicitly map registers to actions. This thesis proposes a novel approach that uses Q-Learning on top of LGP, Reinforced Linear Genetic Programming (RLGP) to learn the optimal register-action assignments. In doing so, we introduce a new framework `linear-gp' written in memory-safe Rust that allows for extensive experimentation for future works.
\end{abstract}

\begin{acknowledgements}
Here's to my young niece Eva, whose curiosity and enthusiasm have inspired me. I hope you find inspiration from my journey and one day embark on your own.

My deepest gratitude goes to my brother Frebarz, who consistently encouraged me to reach new heights.
My mother, Khalima, and my sisters Rita and Rukhafzo have always been my motivation, continuously fueling my ambitions.
I also cherish the memory of my late father, Muhammad Naim Nazar Muhammad (09/25/1958 - 06/29/2006), whose unwavering dedication to his passions has inspired my own journey.

I am incredibly appreciative of my brothers who, although not connected by blood, are nothing short of family.
Their constant encouragement has been invaluable.
A heartfelt thank you goes to Kerel, Colin, Mohammed, Efaj, Assib, Jason, and Chris.
A special thanks goes to all my close friends and colleauges who have enabled to me to grow as a person and as a student.

A massive thank you goes to my highschool teacher, Mr. Oliver Wang, for reigniting my passion for problem-solving.
To Ms. Effie Chan, thank you for introducing me to the world of martial arts, you have forever altered my life for the better.
I am also thankful for Anthony Arena, who persuaded me to take Computer Science and guided me towards a better path when I needed it the most.

Finally, my immense gratitude goes to Dr. Heywood for his patience and unwavering guidance throughout the entire process of this thesis, from its inception to completion. To each and every one of you who have played a significant role in my academic journey, I offer my sincere thanks and appreciation. Your love and support mean the world to me.
\end{acknowledgements}

\mainmatter

\chapter{Introduction}
The rapid growth of technology and the increasing complexity of real-world problems demand efficient and effective optimization techniques. Evolutionary algorithms (EAs) have demonstrated significant potential in addressing these challenges by emulating the process of natural selection to search for optimal solutions \cite{poli08}. Linear Genetic Programming (LGP), a branch of Genetic Programming (GP), is one such EA that has gained attention in the field of computer science for its unique approach to tackling complex problems \cite{song03}. By representing programs as a sequence of linear instructions, similar to what is found when programming imperatively, LGP allows for solutions that are not only potentially more interpretable but also more efficient to execute and manipulate.
The resulting algorithm, Reinforced Linear Genetic Programming (RLGP), is then able to learn the optimal register-action assignments, potentially leading to more efficient and effective solutions. RLGP inherits the benefits of LGP, such as its linear representation and ease of manipulation while augmenting it with the adaptive capabilities of Q-Learning. This integration allows RLGP to explore and exploit the solution space more effectively, resulting in improved performance when addressing complex optimization problems.
We evaluate the performance of RLGP on a variety of benchmark problems, including the cartpole-v1 and mountain-car-v0 environments from the OpenAI Gym library \cite{1606.01540}. These environments provide challenging tasks that require sophisticated decision-making and control strategies, making them suitable for assessing the effectiveness of RLGP. We compare the baseline LGP framework with the augmented RLGP framework in terms of solution quality, convergence speed, and adaptability to dynamic problem domains.
By combining the strengths of LGP and Q-Learning, RLGP might represent a significant advancement in the field of evolutionary computation. This research not only contributes to the understanding of hybrid evolutionary algorithms but also paves the way for future work on incorporating reinforcement learning techniques into other EAs.

\chapter{Background}
\section{Linear Genetic Programming}
Linear Genetic Programming (LGP) is an advanced form of Genetic Programming (GP), a powerful machine learning technique introduced by Brameier and Banzhaf in their seminal work \cite{brameier2001comparison}. Unlike the traditional tree-based GP, LGP represents programs as a linear sequence of instructions, similar to assembly language or machine code \cite{koza93}. This representation offers several advantages, such as improved evolvability, efficient execution, and simplicity of crossover and mutation operations.
In LGP, programs are composed of registers and instructions, where each instruction manipulates the contents of registers using arithmetic, logical, or conditional operations. The evolutionary process is similar to that of canonical GP, but involves variation operators that act directly upon a linear set of instructions. The variations are as follows, reproduction (cloning), recombination (breeding) and mutation.
In their paper, Brameier and Banzhaf compared the performance of LGP to that of traditional GP and neural networks. They discovered that LGP had classification and generalization capabilities that were comparable to those of neural networks \cite{brameier2001comparison}. This finding was significant, as it demonstrated that LGP could serve as an alternative to neural networks for solving complex machine learning problems. Moreover, LGP's linear representation allows for more interpretable solutions, which is an important consideration in many applications where understanding the underlying model is crucial.
The paper further explored the benefits of LGP's linear representation, such as improved evolvability and more efficient execution. These advantages make LGP an attractive choice for solving complex problems, as it can produce high-quality solutions more quickly than traditional GP methods. Additionally, the simplicity of crossover and mutation operations in LGP ensures that the evolutionary process remains efficient and effective, allowing for the exploration of a diverse range of solutions.
Brameier and Banzhaf's work on LGP laid the foundation for further research into the capabilities and applications of this powerful machine learning technique. The findings of their paper highlight the potential of LGP as a robust and versatile machine learning approach, particularly in scenarios where both high performance and interpretability are required.
\section{Reinforced Genetic Programming}
In the RGP paper, Downing applied the Reinforced Genetic Programming approach to several benchmark problems, including function optimization and control tasks \cite{downing1995reinforced}. The results showed that RGP significantly outperformed traditional Genetic Programming and other baseline algorithms in terms of convergence speed, solution quality, and robustness. This demonstrated the effectiveness of incorporating Q-Learning into the genetic programming framework to guide the exploration and exploitation of the search space.
One of the key findings of the paper was that the combination of Genetic Programming and Q-Learning allowed RGP to adapt more efficiently to changing environments and problem landscapes. By utilizing the reinforcement signals from the environment, RGP could dynamically adjust its search strategy, making it more responsive to the changes in the problem domain. This ability to adapt and learn from the environment is particularly relevant to real-world problems, where the solution space may be dynamic, noisy, or uncertain.
The paper also introduced several novel techniques for integrating Q-Learning into the Genetic Programming framework, such as the use of Q-values to bias the selection of genetic operations and the incorporation of reinforcement signals into the fitness function. These innovations allowed RGP to leverage the strengths of both Genetic Programming and Q-Learning, resulting in a more powerful and flexible optimization algorithm.
In the context of Reinforced Linear Genetic Programming (RLGP), the findings of Downing's paper suggest that integrating Q-Learning into the LGP framework could yield similar benefits. By combining the global search capabilities of LGP with the local search and adaptation of Q-Learning, the resulting algorithm, which could be referred to as RLGP, may be able to tackle complex and dynamic problem domains more effectively.
It is important to note that RGP represents programs as decision trees, while LGP uses linear sequences of instructions for program representation. This fundamental difference necessitates adopting a distinct approach when incorporating Q-learning into LGP.
\chapter{Methodology}

\section{Framework Overview}
We utilize the Rust programming language to develop a flexible and extensible framework for our research, available at \url{https://github.com/urmzd/linear-gp}. The framework is specifically designed to address a wide range of problems using LGP and RLGP while offering extensive configurability for experimentation purposes. The framework is composed of multiple engines, each implementing core functionalities essential for the evolutionary process. The primary engines we focus on are described below:

\begin{itemize}
	\item \textbf{Core}: This engine establishes the fundamental processes the framework employs to evolve a population of individuals. It initializes the population, iteratively evaluates the fitness of individuals, performs selection, and applies genetic operations such as mutation and crossover to generate offspring.

	\item \textbf{Breed}: This engine is responsible for defining the genetic operators that facilitate the exchange of genetic material between individuals. In particular, it outlines how individuals undergo crossover, generating offspring with a combination of their parents' genetic code.

	\item \textbf{Mutate}: This engine focuses on the stochastic modification of single individuals within the population. It dictates how genetic alterations are applied to individuals, potentially leading to the discovery of novel and improved solutions.

	\item \textbf{Fitness}: The Fitness engine is tasked with evaluating the performance of each individual in the population. It measures how well an individual can solve the target problem, providing a basis for selection and guiding the evolutionary search process.

	\item \textbf{Generate}: This engine is concerned with the stochastic creation of individuals and environments. It defines the methods for generating new individuals with random genetic material, as well as the procedures for initializing problem-specific environments that influence the evaluation of individuals.
\end{itemize}

\subsection{Framework Configuration \& Hyperparameters}

The available framework configuration properties are listed in \ref{tab:hyperparameters}. Note that note all configurations are always used.
For instance, programs that do not use Q-Learning during their fitness evaluation, $H_{alpha}$, $H_{gamma}$, $H_{epsilon}$, $H_{alpha\_decay}$ or $H_{epsilon\_decay}$ is not required.
\begin{itemize}
	\item $H$ denotes the set of hyperparameters.
	\item $H_{p}$ denotes the set of hyperparameters that are related to a specific property $p$.
\end{itemize}
For example, when referencing gap, we use the notation $H_{gap}$.

\begin{table}[!htb]
	\centering
	\begin{tabular}{|l|l|l|}
		\hline
		\textbf{Property}  & \textbf{Type} & \textbf{Value Range} \\
		\hline
		default\_fitness   & float         & -                    \\
		\hline
		population\_size   & int           & [0,)                 \\
		\hline
		gap                & float         & (0, 1.)              \\
		\hline
		mutation\_percent  & float         & [0, 1.0]             \\
		\hline
		crossover\_percent & float         & [0, 1.0]             \\
		\hline
		n\_generations     & int           & [0,)                 \\
		\hline
		n\_trials          & int           & [1,)                 \\
		\hline
		seed               & int or None   & [0,)                 \\
		\hline
		max\_instructions  & int           & [1,)                 \\
		\hline
		n\_extras          & int           & [1,)                 \\
		\hline
		external\_factor   & float         & [0,)                 \\
		\hline
		n\_actions         & int           & -                    \\
		\hline
		n\_inputs          & int           & -                    \\
		\hline
		alpha              & float         & [0, 1.0]             \\
		\hline
		gamma              & float         & [0, 1.0]             \\
		\hline
		epsilon            & float         & [0, 1]               \\
		\hline
		alpha\_decay       & float         & [0, 1]               \\
		\hline
		epsilon\_decay     & float         & [0, 1]               \\
		\hline
	\end{tabular}
	\caption{Hyperparameters}
	\label{tab:hyperparameters}
\end{table}
The genetic algorithm involves several hyperparameters that influence its behavior. \texttt{default\_fitness} is the default fitness value assigned to individuals in the population when there is an issue with their fitness evaluation. The number of individuals in each generation is determined by \texttt{population\_size}. The \texttt{gap} parameter represents the percentage of the population that is replaced with offspring in each generation. The \texttt{mutation\_percent} and \texttt{crossover\_percent} values control the proportions of individuals generated through mutation and crossover processes, respectively.

The algorithm runs for a set number of generations, specified by \texttt{n\_generations}. During fitness evaluation, a program must interact with an environment a certain number of times, determined by \texttt{n\_trials}, and the results are aggregated to produce a final value. The pseudorandom number generator (PRNG) is initialized with a seed, \texttt{seed}, or noise from the system if set to None. The genetic program has a maximum number of instructions, \texttt{max\_instructions}, and a certain number of working registers, specified by \texttt{n\_extras}.

The \texttt{external\_factor} parameter adjusts the amplification or reduction of inputs from external sources. The program has \texttt{n\_actions} possible choices or action registers, while the number of feature values for an input from an external source is determined by \texttt{n\_inputs}.

The Q-learning algorithm also utilizes various hyperparameters. The initial learning rate, \texttt{alpha}, sets the pace for updating action value estimates. The \texttt{gamma} parameter acts as the discount factor for future rewards, affecting the balance between immediate and long-term rewards. The probability of selecting a random action is controlled by \texttt{epsilon}, allowing the algorithm to explore the state space. Over time, the learning rate and exploration rate decay at rates specified by \texttt{alpha\_decay} and \texttt{epsilon\_decay}, respectively, allowing a shift from exploration to exploitation.

\subsection{Program Representation}
A program can be thought of as a container holding instructions and a set of registers. A single instruction consists of the source register index, the target register index, the operation to be performed and a mode flag. The mode flag is used to determine where the source register is located. If set, the mode flag indicates that the target register is located outside the program. In other words, the value held in the target register is given by an external source (such as the features of some dataset or the environment state). We can represent the program as a sequence of instructions in the following format.
\begin{equation}
	R[y] \leftarrow R[y] \langle op \rangle  R[x]
\end{equation}
$R$ represents the registers, $x$ and $y$ represent the and target register indices respectively, and $\langle op \rangle$ represents the operation to be performed. In our case, we only allow four operations; addition $(+)$,
subtraction $(-)$, multiplication $(\times)$ and division $(\div)$ by $2$. Division is done by a non-zero constant to prevent division by zero errors that would likely crash the system. This representation makes it easy to implement variation operations, but also easy to digest for humans unlike other machine learning techniques and tree-based programs, which convolute the internal process used to solve a problem. The size of the register set consists of $H_{n\_actions} +  H_{n\_extras}$.

\subsection{The Algorithm}
In this work, we present a linear genetic programming (LGP) approach to evolve a population of programs to solve a given task. The core algorithm, as well as supporting operations, are outlined below.
The core LGP algorithm (Algorithm \ref{alg:core}) starts by initializing a population of programs, where each program is evaluated against the desired task using a suitable fitness function. The programs are then ranked based on their fitness scores, and the least fit individuals are dropped by a given percentage, denoted as $H_{gap}$. The remaining individuals in the population are used to produce new offspring, which fill the dropped spots, thus creating a new generation of the population.
The breeding operation, called Two-Point Crossover (Algorithm \ref{alg:breed}), is used to create offspring by combining parts of two parent programs. It starts by cloning the parent programs, selecting two random points in their instruction sets, and swapping the chunks between them. This operation generates two new offspring. However, only one of the offspring is selected at random and returned, as per the modification.
The mutation operation, called Instruction Replacement (Algorithm \ref{alg:mutate}), is used to introduce random changes into a program. It selects a random instruction from the program and replaces it with a newly generated random instruction. Additionally, it may randomly decide to replace only certain properties (operation, source, or target) of the selected instruction.
The program generation operation (Algorithm \ref{alg:generate}) creates a new program by generating a random number of instructions and a register set, consisting of action registers $H_{n\_actions}$ and working registers $H_{n\_extras}$. The resulting program is then returned.
\begin{algorithm}[!htb]
	\caption{Core: Linear Genetic Programming}
	\label{alg:core}
	\begin{algorithmic}[1]
		\State $P_{0} \gets \Call{InitializePopulation}{H_{population\_size}}$
		\For{$i \in \{0, 1, \ldots, H_{n\_generations}-1\}$}
		\State $\Call{Evaluate}{P_{i}}$
		\State $\Call{Rank}{P_{i}}$
		\State $P_{i+1} \gets \Call{Survive}{P_{i}, H_{gap}}$
		\State $P_{i+1} \gets \Call{Variation}{P_{i+1}}$
		\Comment select individuals through tournament selection and apply variation operations
		\EndFor
		\State \Return $P_{H_{n\_generations} - 1}$
	\end{algorithmic}
\end{algorithm}
\begin{algorithm}[!htb]
	\caption{Breed: Two-Point Crossover}
	\label{alg:breed}
	\begin{algorithmic}[1]
		\State{$P_1', P_2' \gets \Call{Clone}{P_1, P_2}$}
		\State{$I_1 \gets P_1'.instructions$}
		\State{$I_2 \gets P_2'.instructions$}
		\State{$p_1, p_2 \gets \Call{RandomChunk}{I_1, I_2}$}
		\State{$I_1[p_1:p_2], I_2[p_1:p_2] \gets I_2[p_1:p_2], I_1[p_1:p_2]$}
		\State \textbf{return} $\Call{RandomOne}{P_1', P_2'}$
	\end{algorithmic}
\end{algorithm}
\begin{algorithm}[!htb]
	\caption{Mutate: Instruction Replacement}
	\label{alg:mutate}
	\begin{algorithmic}[1]
		\State{$I \gets \Call{RandomInstruction}{P}$}
		\State{$I' \gets \Call{GenerateRandomInstruction}{}$}
		\For{$p \in \{operation, source, target\}$}
		\If{$\Call{Random}{0,1} < 0.5$}
		\State{\Call{ReplaceProperty}{$I, I'$}}
		\EndIf
		\EndFor
		\State{$\Call{ReplaceInstruction}{P, I}$}
		\State \textbf{return} $P$
	\end{algorithmic}
\end{algorithm}
\begin{algorithm}[!htb]
	\caption{Generate: Program Generation}
	\label{alg:generate}
	\begin{algorithmic}[1]
		\State{$n\_instructions \gets \Call{Random}{1, max\_instructions}$}
		\State{$R \gets \Call{GenerateRegisterSet}{H_{n\_actions} + H_{n\_extras}}$}
		\State{$I \gets \Call{GenerateInstructions}{n\_instructions}$}
		\State{$P \gets \Call{CreateProgram}{R, I}$}
		\State \textbf{return} $P$
	\end{algorithmic}
\end{algorithm}

\subsection{Validating The Algorithm}
\begin{figure}[!htb]
	\centering
	\begin{subfigure}{1.0\textwidth}
		\includegraphics[width=\linewidth]{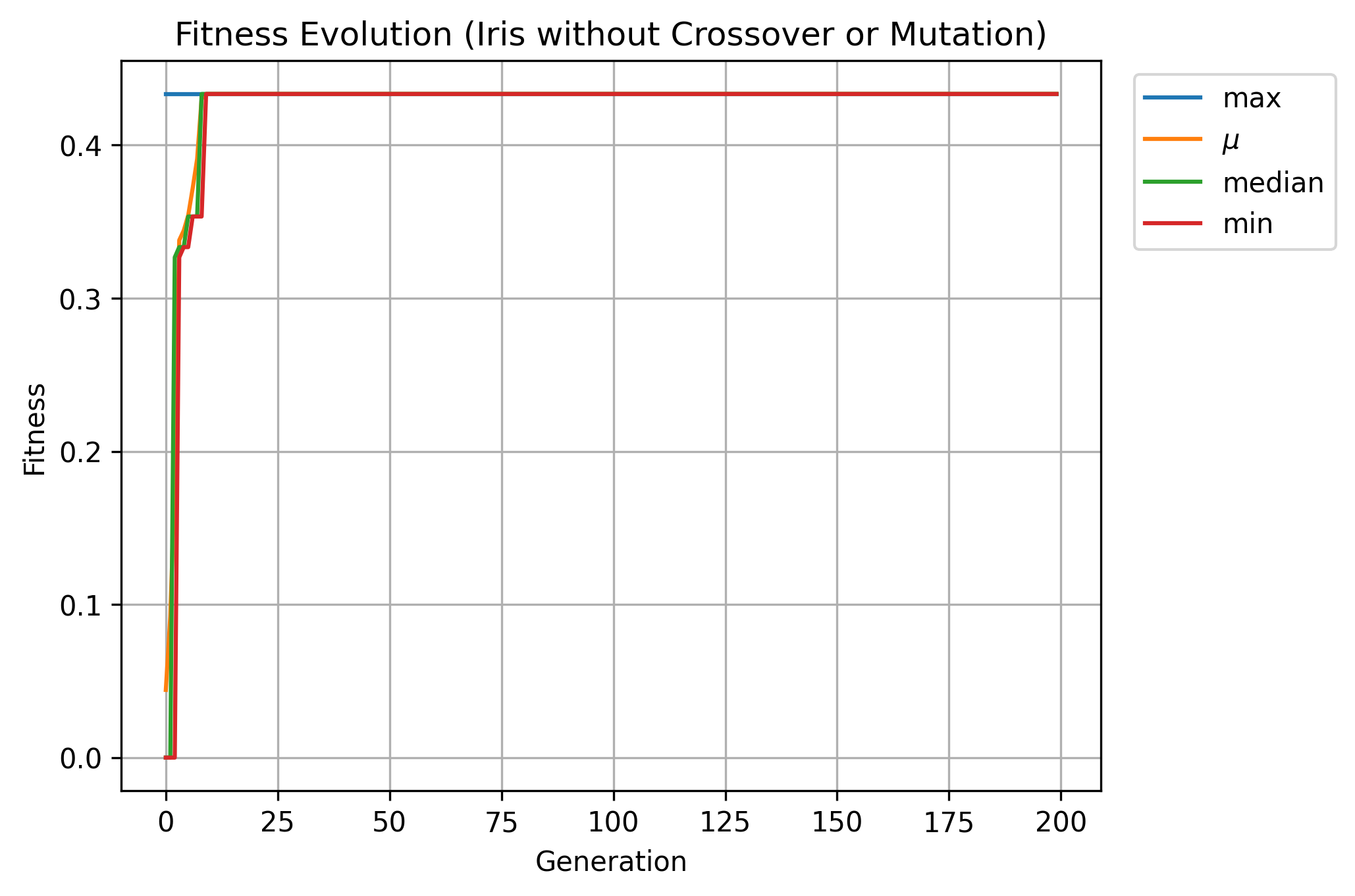}
		\caption{LGP: Iris Reproduction}
		\label{fig:iris_baseline}
	\end{subfigure}

	\begin{subfigure}{1.0\textwidth}
		\includegraphics[width=\linewidth]{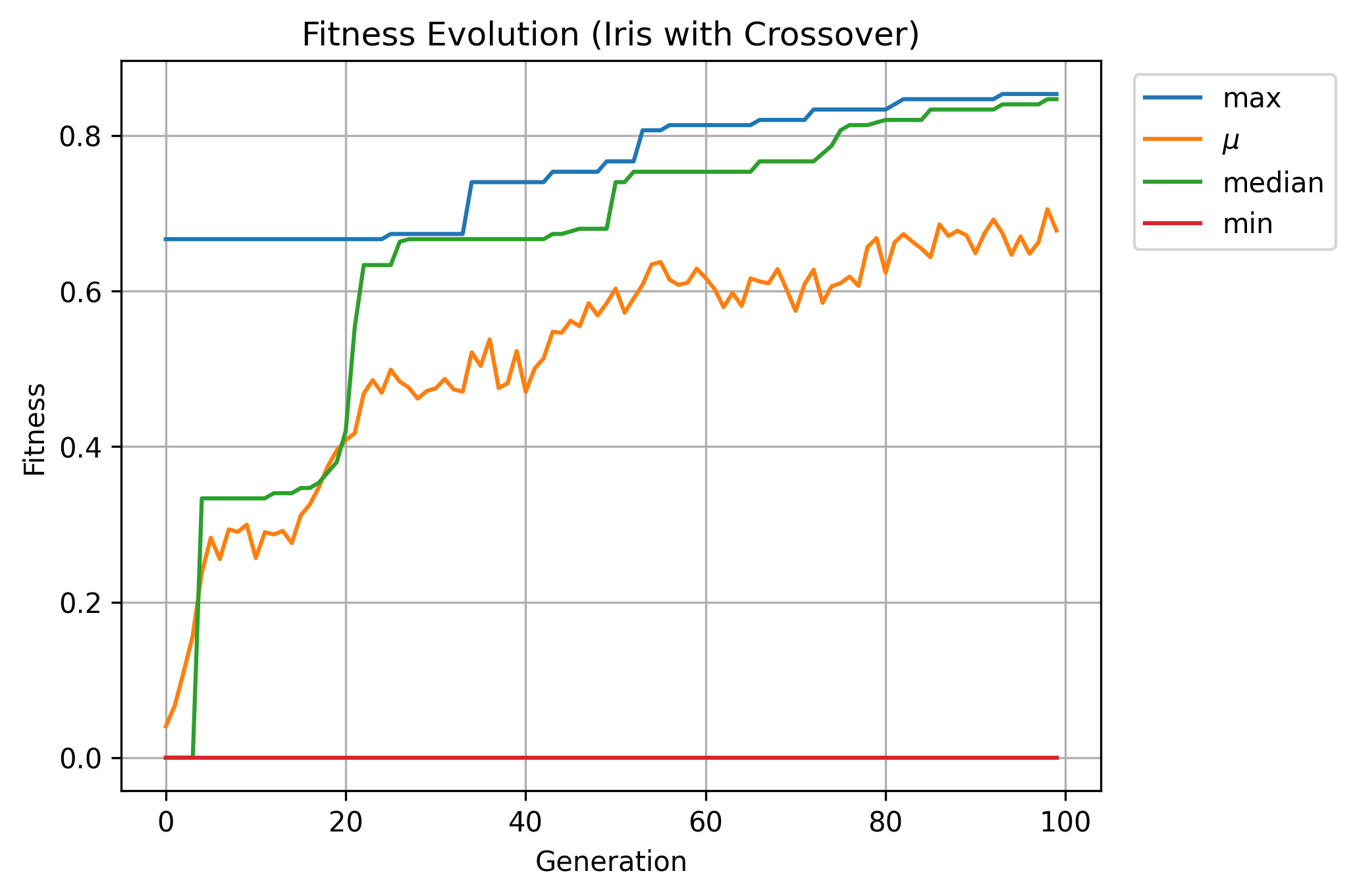}
		\caption{LGP: Iris Recombination}
		\label{fig:iris_crossover}
	\end{subfigure}

	\caption{Performance comparison of different genetic programming methods on iris dataset (1/2)}
	\label{fig:iris_comparison}
\end{figure}

\begin{figure}[!htb]
	\centering
	\begin{subfigure}{1.0\textwidth}
		\includegraphics[width=\linewidth]{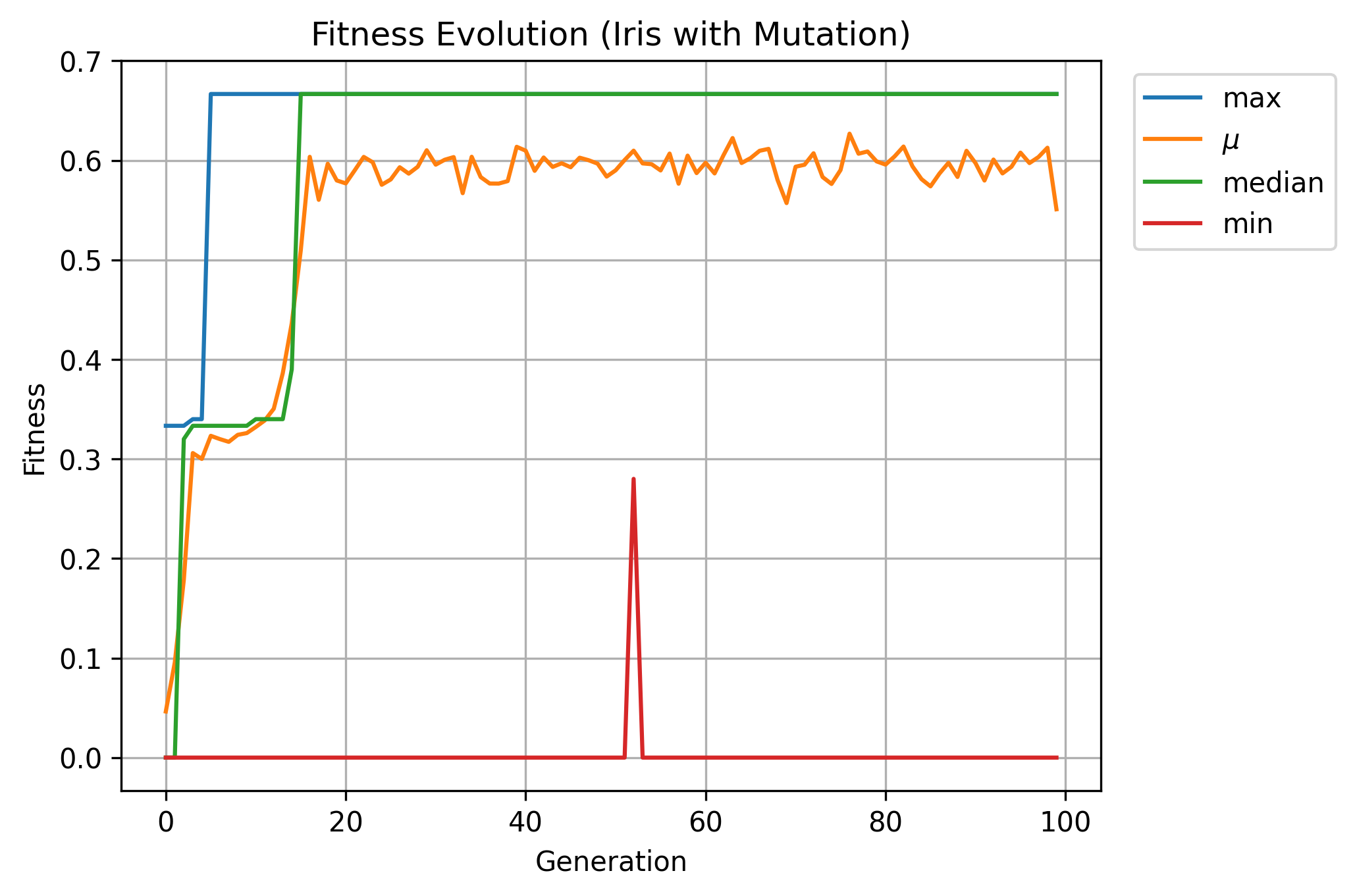}
		\caption{LGP: Iris Mutation}
		\label{fig:iris_mutation}
	\end{subfigure}

	\begin{subfigure}{1.0\textwidth}
		\includegraphics[width=\linewidth]{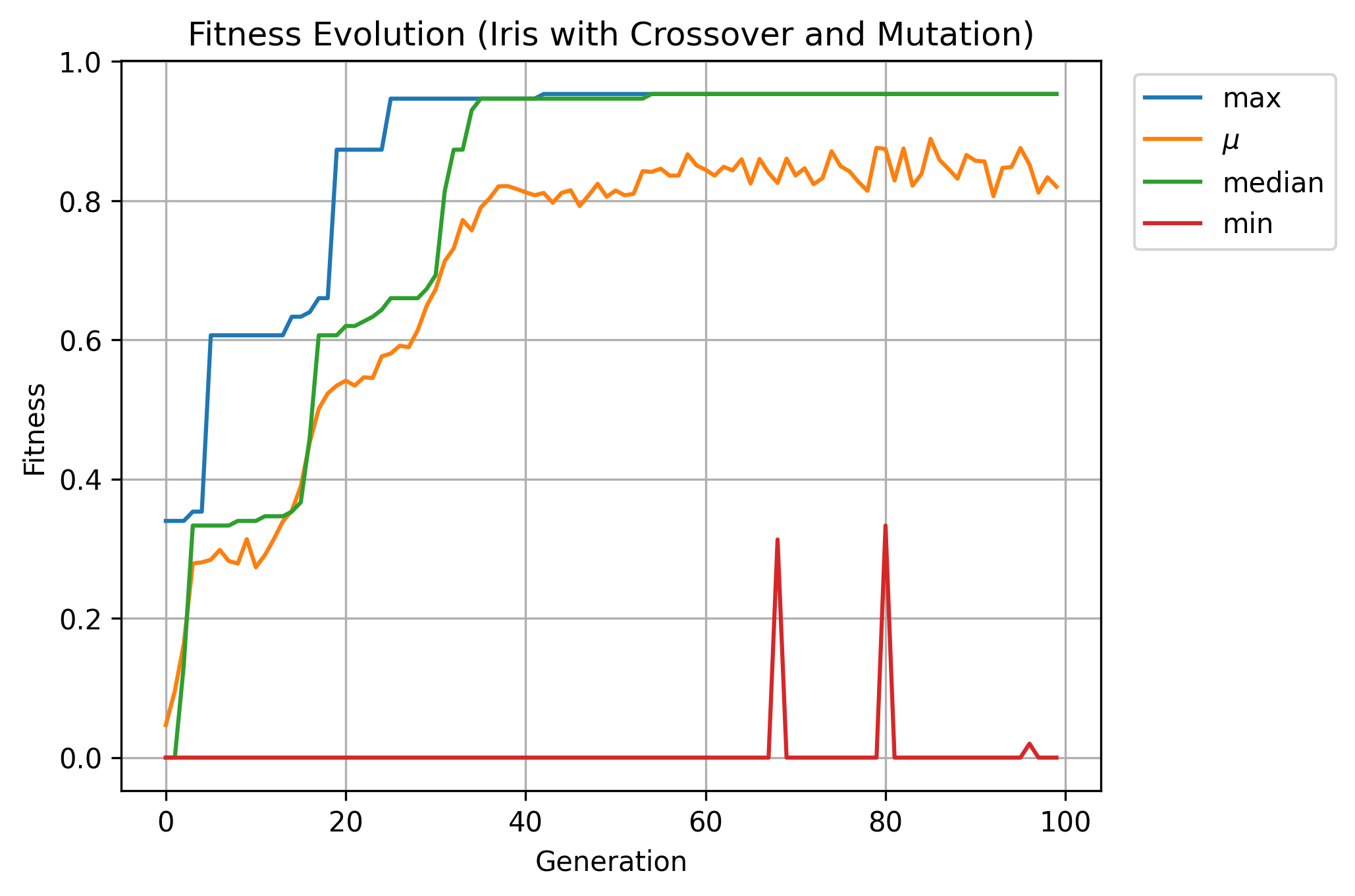}
		\caption{LGP: Iris Recombination \& Mutation}
		\label{fig:iris_full}
	\end{subfigure}
	\caption{Performance comparison of different genetic programming methods on iris dataset (2/2)}
	\label{fig:iris_comparison_continued}
\end{figure}

As a means of ensuring that the framework was implemented correctly, we developed four tests.
Figure \ref{fig:iris_baseline} is a baseline that ensures that the genetic algorithm works with only the reproduction operation.
Figure \ref{fig:iris_crossover}, ensures the two point crossover operation works as expected.
Figure \ref{fig:iris_mutation} ensures that the mutation operation works.
Figure \ref{fig:iris_full} ensures that recombination and mutation can work together to produce a better fitness score than either operation alone.
We also outline the fitness algorithm used for the iris dataset in Algorithm \ref{alg:fitness-iris}.
\begin{algorithm}[!htb]
	\caption{Fitness: Iris}
	\label{alg:fitness-iris}
	\begin{algorithmic}[1]
		\State{$score \gets 0$}
		\For{$I \in inputs$}
		\State{$\Call{Execute}{Program, I}$}
		\State{$pred \gets \Call{Argmax}{Program.Registers[0:H_{n\_actions}]}$}
		\If{$pred = I$}
		\State{$score \pm 1$}
		\EndIf
		\EndFor
		\State{$accuracy \gets \frac{score}{|inputs|}$}
		\State return $accuracy$
	\end{algorithmic}
\end{algorithm}

\section{OpenAI Gym Integration}
We extend the fitness algorithm to work with a Rust port  of OpenAI's mountain-car-v0 and cartpole-v1 environment \cite{1606.01540}. The code can be viewed at (\url{https://github.com/urmzd/gym-rs}).

For the mountain-car-v0 environment, the agent's goal is to drive a car up a steep hill. The car is subject to gravity and has limited power. The car must learn to rock back and forth to build momentum before reaching the goal area at the top of the hill. The environment has two state variables representing the position, and velocity of the car. The car has three possible actions: push left, push right, or no push. The reward function for the environment gives a value of $-1$ at each time step until the car reaches the goal area, at which point the reward becomes $0$. The problem is considered solved when the car reaches the goal area with an average reward of -110 over 100 consecutive trials.

For the cartpole-v1 environment, the agent's goal is to balance a pole on top of a cart that can move left or right. The pole is subject to gravity, and the goal is to keep the pole upright for as long as possible. The environment has four continuous state variables representing the position and velocity of the cart and pole. The cart has two possible actions: move left or move right. The reward function for the environment gives a value of $1$ at each time step that the pole remains upright. The problem is considered solved when the pole remains upright for at least $195$ consecutive time steps over 100 consecutive trials.

With this in mind, we outline an alternative fitness algorithm used to baseline the framework on classical reinforcement learning problems, Algorithm \ref{alg:fitness-gym}.

\begin{algorithm}[!htb]
	\caption{Fitness: Gym Integration}
	\label{alg:fitness-gym}
	\begin{algorithmic}[1]
		\State{$score \gets 0$}
		\For{$i\in N_{Episodes}$}
		\State{$\Call{Execute}{Program, Environment}$}
		\State{$Action \gets \Call{Argmax}{Program.Registers[0:H_{n\_actions}]}$}
		\State{$Reward, Terminal \gets Sim(Action, Environment)$}
		\If{Terminal}
		\State break
		\EndIf
		\State{$score \pm reward$}
		\EndFor
		\State \textbf{return} $score$
	\end{algorithmic}
\end{algorithm}

\section{Q Learning Integration}
We extend Algorithm \ref{alg:fitness-gym} further to support Q Learning (Algorithm \ref{alg:fitness-q}).
The Q Table consists of a 2D array of size $N_R \times N_A$, where $N_R$ is the number of registers the program can work with and $N_A$ is the number of available actions an agent can take. The Q Table is initialized to all zeros and updates only a different register has been selected. The Q Table is updated using the following formula.
\begin{equation}
	Q(R[x_t], a_t) \leftarrow Q(R[x_t], a_t) + \alpha \left(r_{t+1} + \gamma \max_a Q(R[x_{t+1}], a) - Q(R[x_t], a_t)\right)
\end{equation}
In this formula, $Q$ represents the Q Table, $R$ represents the set of registers, $a$ represents the available actions, $r_{t+1}$ is the reward at time step $t+1$, $\alpha$ is the learning rate, and $\gamma$ is the discount factor. The value $x_t$ represents the current state, while $x_{t+1}$ represents the next state. The update rule calculates the new Q value for the current state-action pair based on the observed reward and the maximum Q value for the next state.
Moreover, during register selection, we apply a greedy selection policy (Algorithm \ref{alg:q-learning-greedy-selection}) with probability $1-\epsilon$. $\epsilon$ exists to allow us to configure the exploration-exploitation factor, i.e., maintain register selection diversity.

\begin{algorithm}[!htb]
	\caption{Q Learning: $\epsilon$-Greedy Selection Policy}
	\label{alg:q-learning-greedy-selection}
	\begin{algorithmic}[1]
		\State $\Call{Execute}{Program, State}$
		\State $WinningRegister \gets \Call{Argmax}{Program.Registers} $
		\State $Action \gets \Call{Argmax}{Q[WinningRegister]}$
		\If{$\Call{Random}{0, 1} < \epsilon$}
		\State $Action \gets \Call{Random}{0, H_{n\_actions}}$
		\EndIf
		\State \textbf{return} $WinningRegister, Action$
	\end{algorithmic}
\end{algorithm}
\begin{algorithm}[!htb]
	\caption{Fitness: Q Learning}
	\label{alg:fitness-q}
	\begin{algorithmic}[1]
		\State $score \gets 0$
		\State $R_{t}, a_{t} \gets \Call{GreedySelection}{Program, Environment}$
		\While{$t < N_{episodes}$}
		\State $r_{t + 1}, Terminal \gets \Call{Sim}{a_{t}, Environment}$
		\State $score \pm r_{t + 1}$
		\If{$\Call{Terminal}{}$}
		\State \textbf{break}
		\EndIf
		\State $R_{t+1}, a_{t+1} \gets \Call{GreedySelection}{Program, Environment}$
		\If{$R_{t} \neq R_{t+1}$}
		\State $\Call{QTableUpdate}{R_{t+1}, a_{t+1}, R_{t}}$
		\EndIf
		\State ${R_{t} \leftarrow R_{t+1}}$
		\State ${a_{t} \leftarrow a_{t+1}}$
		\State ${t \leftarrow t+1}$
		\EndWhile
		\State \Return $score$
	\end{algorithmic}
\end{algorithm}

\section{Experiment Setup}

In this section, we outline the experiment setup used to evaluate the performance of the proposed Q-Learning LGP algorithm in comparison to the baseline LGP. The experiments were conducted in two main stages: hyperparameter optimization and performance evaluation.

\subsection{Hyperparameter Optimization}
First, we employed the Optuna hyperparameter optimization library \cite{akiba2019optuna} to identify the optimal parameters for the baseline LGP programs.
Optuna is a flexible and efficient hyperparameter optimization framework that enables automatic exploration of hyperparameter spaces in pursuit of the best settings for a given algorithm.
After determining the optimal parameters for the baseline LGP programs, we conducted a separate search using Optuna to find the optimal Q-Learning constants for the Reinforced Linear Genetic Programming (RLGP) programs.
This process was aimed at fine-tuning the RLGP algorithm's performance by selecting the most suitable Q-Learning parameters based on the problem domain and the characteristics of the LGP framework.
The resulting parameters were then integrated into the RLGP algorithm by extending the LGP parameters, as shown in \ref{code:cart-pole-parameters}, \ref{code:cart-pole-q-parameters}, \ref{code:mountain-car-parameters}, and \ref{code:mountain-car-q-parameters}.

\begin{figure}[!htb]
	\centering
\begin{lstlisting}[language={}]
{
    "default_fitness": 500.0,
    "population_size": 100,
    "gap": 0.5,
    "mutation_percent": 0.5,
    "crossover_percent": 0.5,
    "n_generations": 100,
    "n_trials": 100,
    "seed": null,
    "program_parameters": {
        "max_instructions": 23,
        "instruction_generator_parameters": {
            "n_extras": 1,
            "external_factor": 92.04438205753976,
            "n_actions": 2,
            "n_inputs": 4
        }
    }
}
\end{lstlisting}
	\caption{Cart Pole Parameters}
	\label{code:cart-pole-parameters}
\end{figure}
\begin{figure}[!htb]
	\centering
\begin{lstlisting}[language={}]
{
    "default_fitness": 500.0,
    "population_size": 100,
    "gap": 0.5,
    "mutation_percent": 0.5,
    "crossover_percent": 0.5,
    "n_generations": 100,
    "n_trials": 100,
    "seed": null,
    "program_parameters": {
        "program_parameters": {
            "max_instructions": 23,
            "instruction_generator_parameters": {
                "n_extras": 1,
                "external_factor": 92.04438205753976,
                "n_actions": 2,
                "n_inputs": 4
            }
        },
        "consts": {
            "alpha": 0.9933093715472482,
            "gamma": 0.9493877958652062,
            "epsilon": 0.7024493518448414,
            "alpha_decay": 0.24276313855515808,
            "epsilon_decay": 0.293833697874351
        }
    }
}
\end{lstlisting}
	\caption{Cart Pole Q-Learning Parameters}
	\label{code:cart-pole-q-parameters}
\end{figure}
\begin{figure}[!htb]
	\centering
\begin{lstlisting}[language={}]
{
    "default_fitness": -200.0,
    "population_size": 100,
    "gap": 0.5,
    "mutation_percent": 0.5,
    "crossover_percent": 0.5,
    "n_generations": 100,
    "n_trials": 5,
    "seed": null,
    "program_parameters": {
        "max_instructions": 10,
        "instruction_generator_parameters": {
            "n_extras": 1,
            "external_factor": 2.0736207078591695,
            "n_actions": 3,
            "n_inputs": 2
        }
    }
}
\end{lstlisting}
	\caption{Mountain Car Parameters}
	\label{code:mountain-car-parameters}
\end{figure}
\begin{figure}[!htb]
	\centering
\begin{lstlisting}[language={}]
{
    "default_fitness": -200.0,
    "population_size": 100,
    "gap": 0.5,
    "mutation_percent": 0.5,
    "crossover_percent": 0.5,
    "n_generations": 100,
    "n_trials": 100,
    "seed": null,
    "program_parameters": {
        "program_parameters": {
            "max_instructions": 10,
            "instruction_generator_parameters": {
                "n_extras": 1,
                "external_factor": 2.0736207078591695,
                "n_actions": 3,
                "n_inputs": 2
            }
        },
        "consts": {
            "alpha": 0.9973629496495072,
            "gamma": 0.39901321062297757,
            "epsilon": 0.8400771173101154,
            "alpha_decay": 0.6876951222663272,
            "epsilon_decay": 0.5125287069666674
        }
    }
}
\end{lstlisting}
	\caption{Mountain Car Q-Learning Parameters}
	\label{code:mountain-car-q-parameters}
\end{figure}

\subsection{Performance Evaluation}

After obtaining the optimal parameters for both baseline LGP and RLGP, we proceeded with the performance evaluation phase. This stage involved conducting 100 experiments, each consisting of 100 trials. The goal of these experiments was to assess the effectiveness and robustness of the RLGP algorithm compared to the baseline LGP across multiple runs and problem instances.

For each experiment, we calculated the mean, median, minimum, and maximum performance scores obtained by both baseline LGP and RLGP. These summary statistics provided a comprehensive overview of the algorithms' performance, highlighting their strengths and weaknesses across different trials and problem instances.

Finally, we plot the average mean, median, min, max for each experiment to visually compare the performance of the baseline LGP and RLGP algorithms. This visualization allowed us to identify trends and patterns in the algorithms' performance and gain insights into the benefits of incorporating Q-Learning into the LGP framework.

By following this experimental setup, we aimed to provide a thorough and unbiased evaluation of the proposed Q-Learning LGP algorithm, demonstrating its potential advantages over the baseline LGP and paving the way for future research and development in this area.

\chapter{Analysis}

This chapter analyzes the performance of a population of programs trained using Linear Genetic Programming (LGP)
and a population of programs trained using Linear Genetic Programming Reinforced with Q-Learning (LGP-Q). The programs
are trained on the cartpole-v1 and mountain-car-v0 environments \cite{1606.01540} as mentioned in the Reinforcement Integration section.
Once the results are explained, we analyze them to provide insight into the impact of Q-Learning on the performance of LGP. Note that we use the definition of done as referenced on \url{https://github.com/openai/gym/wiki/Leaderboard}.

\section{Experimental Results}

\subsection{Cart Pole}

\begin{figure}[b]
	\centering
	\begin{subfigure}{1.0\textwidth}
		\includegraphics[width=\linewidth]{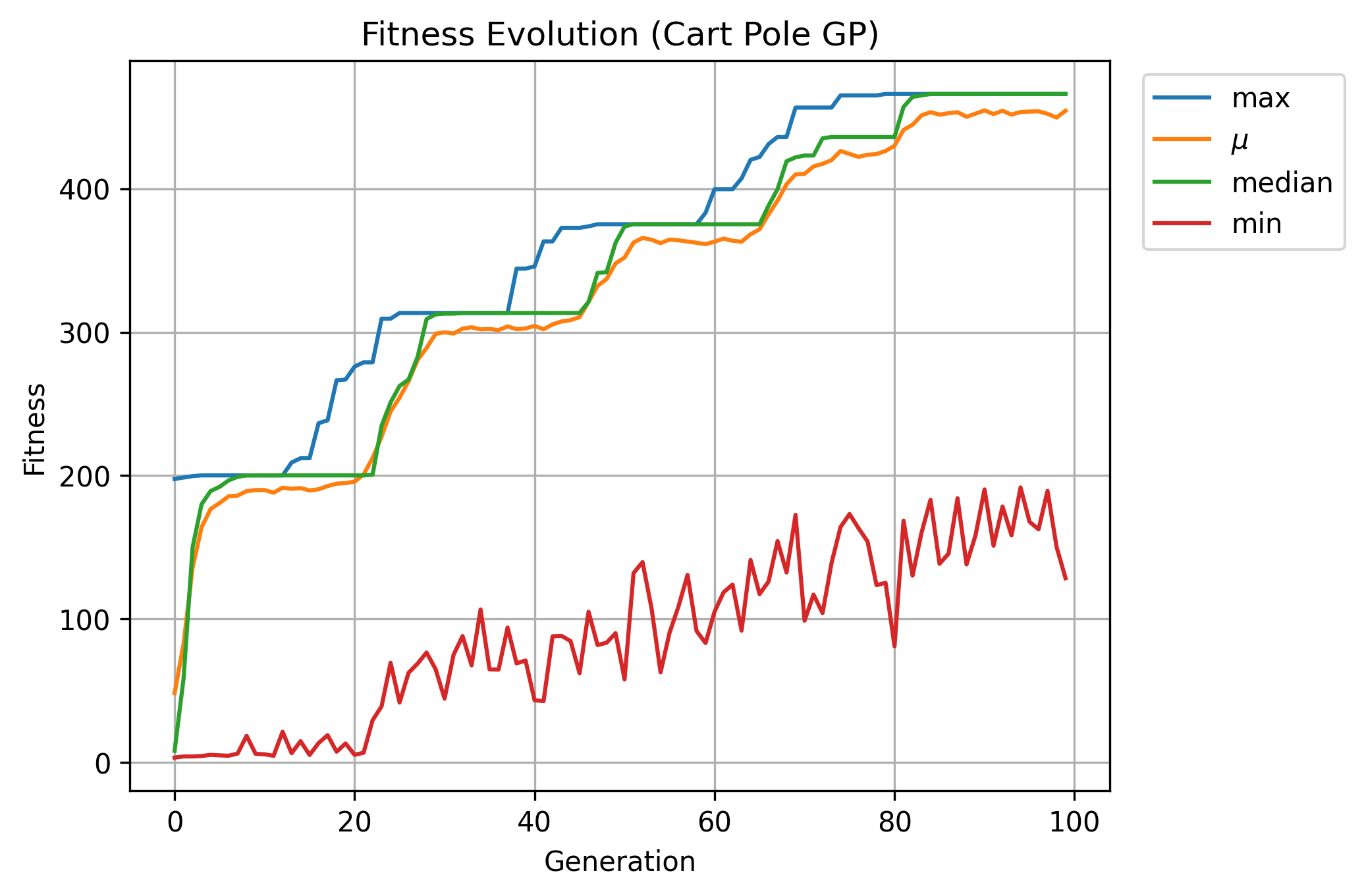}
		\caption{Performance of LGP on Cart Pole}
		\label{fig:cart-pole-lgp}
	\end{subfigure}
	\hfill
	\begin{subfigure}{1.0\textwidth}
		\includegraphics[width=\linewidth]{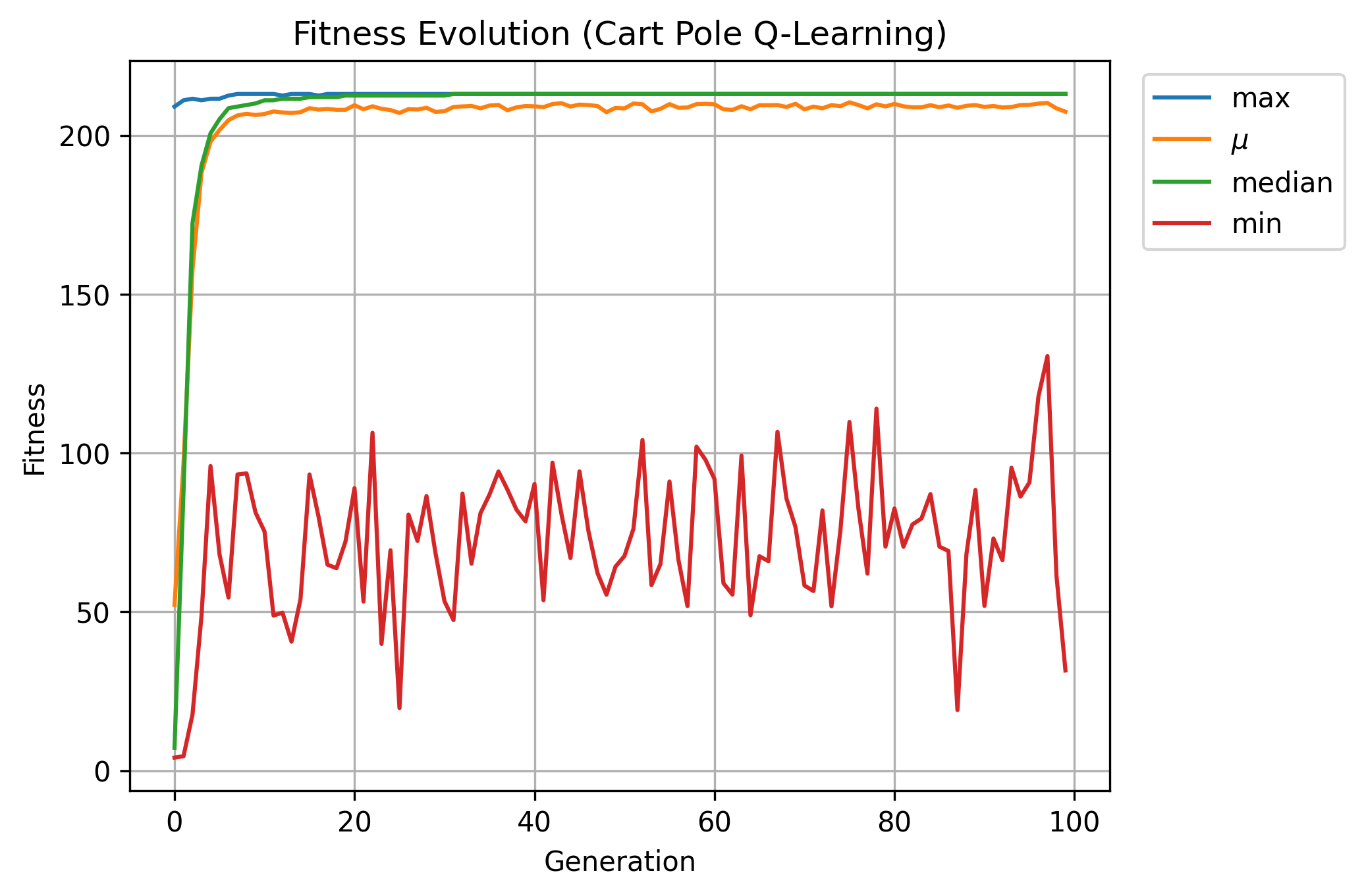}
		\caption{Performance of RLGP on Cart Pole}
		\label{fig:cart-pole-q}
	\end{subfigure}
	\caption{Comparison of RLGP vs LGP Performance on Cart Pole}
	\label{fig:cart-pole-comparison}
\end{figure}

Figure \ref{fig:cart-pole-comparison} with reference tables \ref{tab:cart-pole-lgp} and \ref{tab:cart-pole-q} show a comparison of the two frameworks performance on the cartpole-v0 task.
Over 100 experiments, we observe that LGP averages a maximum of 466, a median of 454, a median of 466 and minimum of 128. On the other hand, RLGP
averages a maximum of 213, a median of 207, a mean of 213 a minimum of 31. In both cases,the population of the respective frameworks were able solve the problem. Both frameworks were able
to generate a program in the first 10 generations that was able to solve the problem. However, the RLGP framework immediately plateaus, whilst the LGP framework continues to improve until hitting a plateau at 80 generations.

\subsection{Mountain Car}

\begin{figure}[b]
	\centering
	\begin{subfigure}{1.0\textwidth}
		\includegraphics[width=\linewidth]{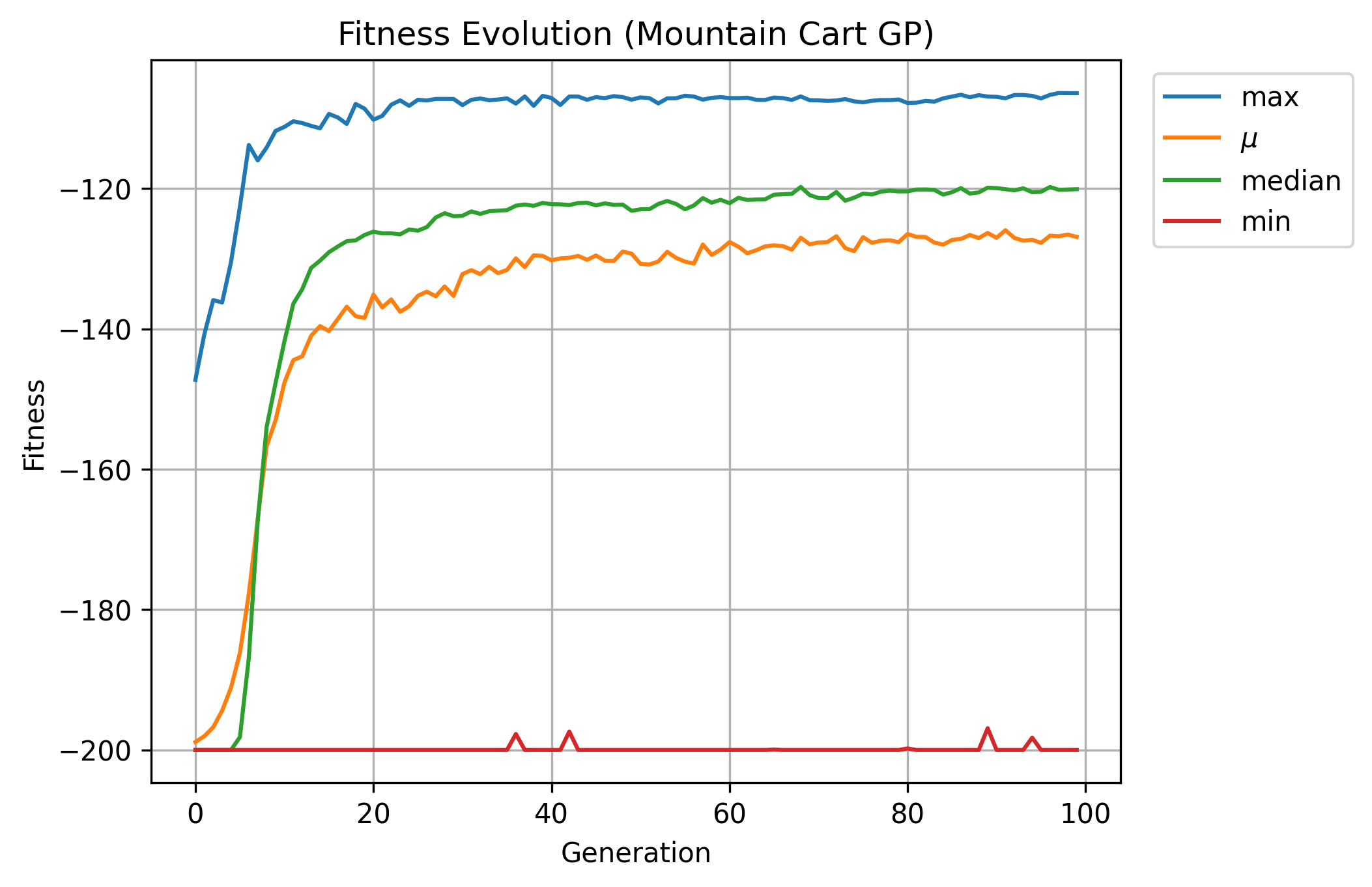}
		\caption{Using LGP on Mountain Car}
		\label{fig:mountain-car-lgp}
	\end{subfigure}
	\hfill
	\begin{subfigure}{1.0\textwidth}
		\includegraphics[width=\linewidth]{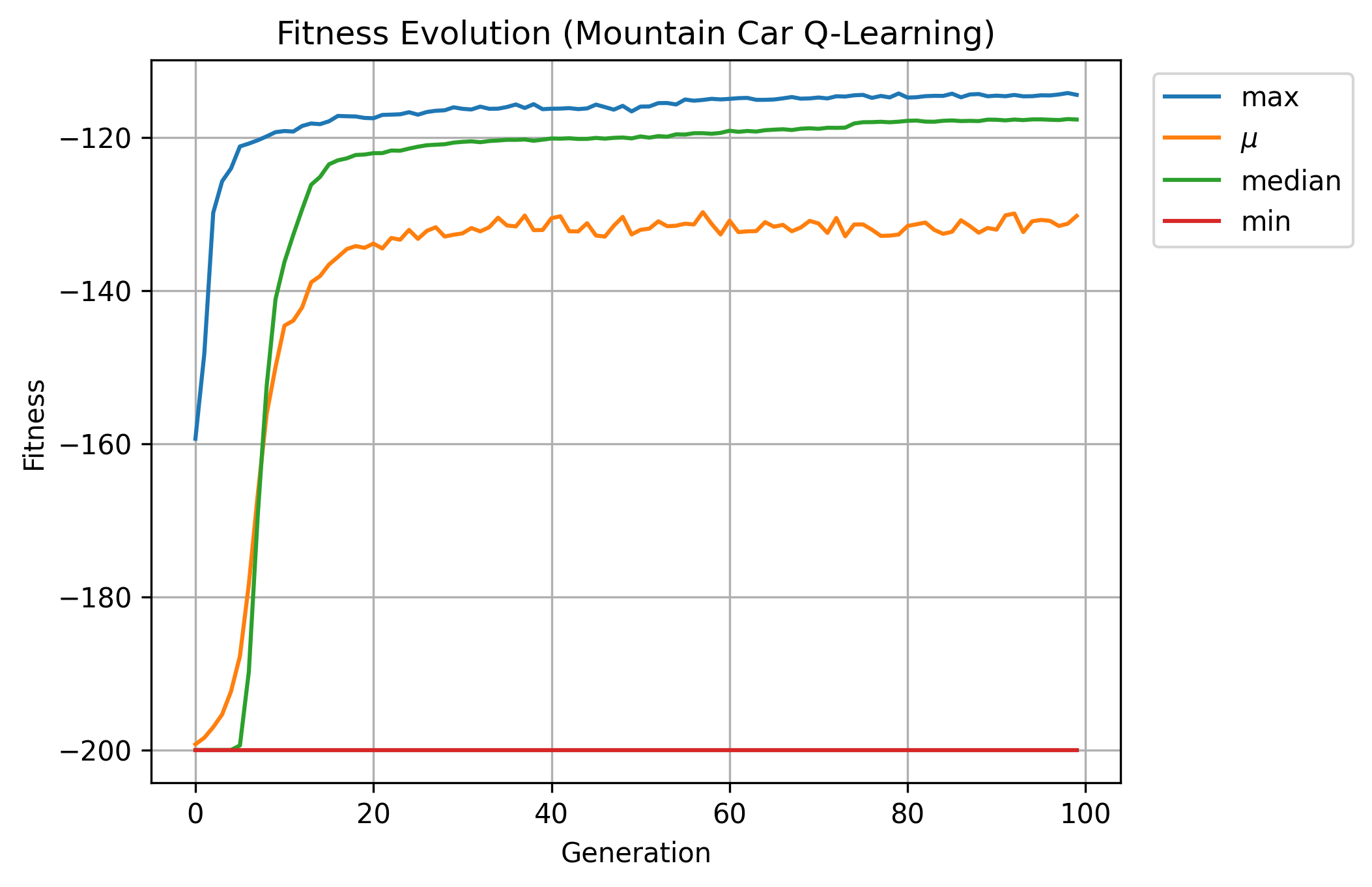}
		\caption{Using RLGP on Mountain Car}
		\label{fig:mountain-car-q}
	\end{subfigure}
	\caption{Comparison of RLGP vs LGP performance on Mountain Car}
	\label{fig:mountain-car-comparison}
\end{figure}

Figure \ref{fig:mountain-car-comparison} alongside tables \ref{tab:mountain-car-lgp} and \ref{tab:mountain-car-q} demonstrates a similar pattern to that of the cartpole-v0 task. The RLGP framework falls short of solving the task, averaging a maximum of -114, a mean of-130, a median of -117 and a minimum of -200. On the other hand, the LGP framework is able to solve the task, albeit narrowly, averaging a maximum -106, a mean of -126, a median of -120,and a minimum -200. Here, we observe that RLGP achieves a higher median score, with a slight upward trend whereas the LGP achieves a lower median, with no particular trend in any direction. Both of these framework look as they quickly converge to a value and then plateau.

\section{Discussion}

Initially, we would have expected for the frameworks to perform better on mountain-car-v0 than cart-pole-v1 as the state space is much simpler (one consists of 2 state properties whilst the other contains four). However, the opposite seems to be true,
the cart pole task was able to be solved relatively easily in comparison to the mountain car task, in which RLGP failed to solve the problem. One explanation for this discrepancy is the difference in the nature of the tasks. The cart-pole task is inherently more continuous, allowing for small adjustments to have a direct impact on the balancing of the pole. On the other hand, the mountain car task requires more strategic and discrete actions to build up momentum and reach the goal. This may indicate that the LGP and RLGP frameworks are better suited to tackle continuous control problems.

Another contributing factor could be the exploration-exploitation trade-off present in RLGP. The Q-learning component might not be exploring the state space effectively, thus getting stuck in local optima and hindering the overall performance. In contrast, the LGP framework seems to be more robust in its exploration of the solution space, enabling it to perform better on both tasks.

Moreover, the difference in performance may also be attributed to the limitations of the genetic programming approach. While LGP and RLGP are capable of discovering compact and interpretable representations of policies, they are not guaranteed to find the global optimum. The search process depends on the initial population and variation operators, which can impact the quality of the solutions found. The experimental set up might have been poor, and the parameters beneficial to LGP might not have been optimal for RLGP. Its possible that
we could've seen better results if Q learning parameters was not simply a wrapper on top of a preexisting configuration.

In conclusion, our experiments demonstrate that the LGP framework outperforms the RLGP framework in both cart-pole and mountain car tasks. Although both frameworks were able to solve the cart-pole task, RLGP failed to solve the mountain car task, suggesting that the integration of Q-learning into genetic programming may not always lead to improved performance. Future work could involve investigating alternative reinforcement learning algorithms, adapting exploration strategies, or incorporating domain knowledge to enhance the performance of genetic programming-based frameworks. Additionally, it would be valuable to test these frameworks on a broader range of tasks to better understand their strengths and limitations.

\resultstable{LGP Cart Pole Aggregated Results}{tab:cart-pole-lgp}{%
0 & 197.6150 & 48.2211 & 8.0180 & 3.3220 \\\hline
1 & 198.6100 & 82.5966 & 58.6020 & 4.1180 \\\hline
2 & 199.6040 & 135.8412 & 149.9540 & 4.1400 \\\hline
3 & 200.1010 & 163.9173 & 180.0105 & 4.4240 \\\hline
4 & 200.1010 & 176.6355 & 189.1775 & 5.2090 \\\hline
5 & 200.1010 & 180.8190 & 192.1495 & 4.8770 \\\hline
6 & 200.1010 & 185.5329 & 196.6230 & 4.6130 \\\hline
7 & 200.1010 & 186.0293 & 199.1070 & 6.1150 \\\hline
8 & 200.1010 & 189.1239 & 200.1010 & 18.4880 \\\hline
9 & 200.1010 & 189.9061 & 200.1010 & 5.9680 \\\hline
10 & 200.1010 & 189.9119 & 200.1010 & 5.6250 \\\hline
11 & 200.1010 & 188.0337 & 200.1010 & 4.5700 \\\hline
12 & 200.1010 & 191.5232 & 200.1010 & 21.3280 \\\hline
13 & 209.0830 & 190.7467 & 200.1010 & 6.4070 \\\hline
14 & 212.0770 & 191.2533 & 200.1010 & 14.7780 \\\hline
15 & 212.0770 & 189.6393 & 200.1010 & 5.1520 \\\hline
16 & 236.5280 & 190.3997 & 200.1010 & 13.4380 \\\hline
17 & 238.5240 & 192.6686 & 200.1010 & 18.9550 \\\hline
18 & 266.4680 & 194.3270 & 200.1010 & 7.4430 \\\hline
19 & 266.9670 & 194.7333 & 200.1010 & 13.1660 \\\hline
20 & 275.9490 & 195.7917 & 200.1010 & 5.2860 \\\hline
21 & 278.9430 & 200.7326 & 200.1010 & 6.6770 \\\hline
22 & 278.9430 & 212.2020 & 200.6000 & 29.3420 \\\hline
23 & 309.3820 & 227.0722 & 234.7815 & 39.0850 \\\hline
24 & 309.3820 & 244.6299 & 250.9990 & 69.4160 \\\hline
25 & 313.3740 & 254.3056 & 262.4760 & 41.6980 \\\hline
26 & 313.3740 & 265.7745 & 266.9670 & 62.4860 \\\hline
27 & 313.3740 & 280.5459 & 282.6855 & 68.8720 \\\hline
28 & 313.3740 & 288.9686 & 309.1325 & 76.5810 \\\hline
29 & 313.3740 & 298.8313 & 312.3760 & 65.0090 \\\hline
30 & 313.3740 & 299.8838 & 312.8750 & 44.4450 \\\hline
31 & 313.3740 & 299.0033 & 312.8750 & 75.0660 \\\hline
32 & 313.3740 & 302.4545 & 313.3740 & 88.0540 \\\hline
33 & 313.3740 & 303.3941 & 313.3740 & 67.6170 \\\hline
34 & 313.3740 & 301.9708 & 313.3740 & 106.6260 \\\hline
35 & 313.3740 & 302.2215 & 313.3740 & 64.8710 \\\hline
36 & 313.3740 & 301.4519 & 313.3740 & 64.6910 \\\hline
37 & 313.3740 & 303.9914 & 313.3740 & 94.0020 \\\hline
38 & 344.3120 & 302.0590 & 313.3740 & 69.0300 \\\hline
39 & 344.3120 & 302.5775 & 313.3740 & 71.0180 \\\hline
40 & 345.8090 & 304.3503 & 313.3740 & 43.3230 \\\hline
41 & 363.2740 & 302.0053 & 313.3740 & 42.6870 \\\hline
42 & 363.2740 & 305.4199 & 313.3740 & 87.9220 \\\hline
43 & 372.7550 & 307.4407 & 313.3740 & 88.1740 \\\hline
44 & 372.7550 & 308.3930 & 313.3740 & 84.6030 \\\hline
45 & 372.7550 & 310.4794 & 313.3740 & 62.1780 \\\hline
46 & 373.7530 & 321.1332 & 321.1085 & 105.0250 \\\hline
47 & 375.2500 & 332.2364 & 341.3180 & 81.8050 \\\hline
48 & 375.2500 & 336.9924 & 341.8170 & 83.4290 \\\hline
49 & 375.2500 & 347.9421 & 362.2760 & 90.0740 \\\hline
50 & 375.2500 & 352.0232 & 373.7530 & 57.8410 \\\hline
51 & 375.2500 & 362.5025 & 375.2500 & 131.9930 \\\hline
52 & 375.2500 & 365.6847 & 375.2500 & 139.6340 \\\hline
53 & 375.2500 & 364.4538 & 375.2500 & 107.3460 \\\hline
54 & 375.2500 & 362.0602 & 375.2500 & 62.7520 \\\hline
55 & 375.2500 & 364.5757 & 375.2500 & 90.4170 \\\hline
56 & 375.2500 & 364.0330 & 375.2500 & 108.8500 \\\hline
57 & 375.2500 & 363.2028 & 375.2500 & 130.7540 \\\hline
58 & 375.2500 & 362.2958 & 375.2500 & 91.5940 \\\hline
59 & 383.2340 & 361.3570 & 375.2500 & 83.2020 \\\hline
60 & 399.7010 & 363.0938 & 375.2500 & 105.0640 \\\hline
61 & 399.7010 & 365.2156 & 375.2500 & 118.5440 \\\hline
62 & 399.7010 & 363.7387 & 375.2500 & 123.9990 \\\hline
63 & 407.1860 & 363.0690 & 375.2500 & 91.8780 \\\hline
64 & 420.1600 & 368.2432 & 375.2500 & 141.1230 \\\hline
65 & 422.1560 & 371.7351 & 375.2500 & 117.3480 \\\hline
66 & 431.1380 & 382.1591 & 388.2240 & 126.0680 \\\hline
67 & 436.1280 & 391.6835 & 399.7010 & 154.2770 \\\hline
68 & 436.1280 & 403.1306 & 419.1620 & 132.3990 \\\hline
69 & 456.5870 & 410.0405 & 421.9065 & 172.5970 \\\hline
70 & 456.5870 & 410.4119 & 423.1540 & 98.7730 \\\hline
71 & 456.5870 & 415.5888 & 423.1540 & 116.9810 \\\hline
72 & 456.5870 & 417.3863 & 435.1300 & 104.1240 \\\hline
73 & 456.5870 & 419.9508 & 436.1280 & 139.0310 \\\hline
74 & 465.0700 & 426.3158 & 436.1280 & 164.1080 \\\hline
75 & 465.0700 & 424.2869 & 436.1280 & 173.1660 \\\hline
76 & 465.0700 & 422.2207 & 436.1280 & 163.2670 \\\hline
77 & 465.0700 & 423.6959 & 436.1280 & 154.0330 \\\hline
78 & 465.0700 & 424.1987 & 436.1280 & 123.5780 \\\hline
79 & 466.0680 & 426.2938 & 436.1280 & 125.2400 \\\hline
80 & 466.0680 & 429.9999 & 436.1280 & 80.9490 \\\hline
81 & 466.0680 & 441.0242 & 457.0860 & 168.5490 \\\hline
82 & 466.0680 & 444.6226 & 464.0720 & 130.1640 \\\hline
83 & 466.0680 & 451.1316 & 465.0700 & 160.2080 \\\hline
84 & 466.0680 & 453.3817 & 466.0680 & 183.1190 \\\hline
85 & 466.0680 & 451.7100 & 466.0680 & 138.4610 \\\hline
86 & 466.0680 & 452.6550 & 466.0680 & 145.5250 \\\hline
87 & 466.0680 & 453.3730 & 466.0680 & 184.1610 \\\hline
88 & 466.0680 & 450.1951 & 466.0680 & 138.0030 \\\hline
89 & 466.0680 & 452.3601 & 466.0680 & 158.3680 \\\hline
90 & 466.0680 & 454.6304 & 466.0680 & 190.3290 \\\hline
91 & 466.0680 & 452.0814 & 466.0680 & 151.0820 \\\hline
92 & 466.0680 & 454.3676 & 466.0680 & 178.3690 \\\hline
93 & 466.0680 & 451.6843 & 466.0680 & 158.2490 \\\hline
94 & 466.0680 & 453.4915 & 466.0680 & 191.7260 \\\hline
95 & 466.0680 & 453.7976 & 466.0680 & 167.7830 \\\hline
96 & 466.0680 & 453.9447 & 466.0680 & 162.4970 \\\hline
97 & 466.0680 & 452.2639 & 466.0680 & 189.3070 \\\hline
98 & 466.0680 & 449.5967 & 466.0680 & 150.4170 \\\hline
99 & 466.0680 & 454.3901 & 466.0680 & 128.4370 \\\hline
}

\resultstable{RLGP Cart Pole Aggregated Results}{tab:cart-pole-q}{%
0 & 209.1090 & 52.2117 & 7.2245 & 4.1240 \\\hline
1 & 211.0950 & 96.6963 & 87.9155 & 4.5420 \\\hline
2 & 211.5910 & 157.7400 & 172.4700 & 17.6940 \\\hline
3 & 211.0930 & 188.2812 & 190.5340 & 48.7950 \\\hline
4 & 211.5870 & 198.0718 & 200.6880 & 95.9330 \\\hline
5 & 211.5870 & 201.7260 & 205.1595 & 68.1670 \\\hline
6 & 212.5820 & 204.8469 & 208.6110 & 54.5070 \\\hline
7 & 213.0780 & 206.3272 & 209.1160 & 93.2930 \\\hline
8 & 213.0780 & 206.8498 & 209.6090 & 93.6100 \\\hline
9 & 213.0780 & 206.4557 & 210.1020 & 81.2950 \\\hline
10 & 213.0780 & 206.7860 & 211.0890 & 75.3410 \\\hline
11 & 213.0780 & 207.5813 & 211.0900 & 48.8610 \\\hline
12 & 212.5810 & 207.2677 & 211.5860 & 49.7480 \\\hline
13 & 213.0780 & 207.0827 & 211.5860 & 40.6320 \\\hline
14 & 213.0780 & 207.3593 & 211.5860 & 53.9910 \\\hline
15 & 213.0780 & 208.6030 & 212.0840 & 93.2840 \\\hline
16 & 212.5810 & 208.1538 & 212.0840 & 80.0530 \\\hline
17 & 213.0780 & 208.3412 & 212.0840 & 64.8700 \\\hline
18 & 213.0780 & 208.1041 & 212.0840 & 63.7540 \\\hline
19 & 213.0780 & 208.1036 & 212.5800 & 72.1530 \\\hline
20 & 213.0780 & 209.5479 & 212.5800 & 88.9900 \\\hline
21 & 213.0780 & 208.2771 & 212.5800 & 53.2580 \\\hline
22 & 213.0780 & 209.2043 & 212.5800 & 106.4120 \\\hline
23 & 213.0780 & 208.4108 & 212.5800 & 39.9400 \\\hline
24 & 213.0780 & 208.0172 & 212.5800 & 69.3860 \\\hline
25 & 213.0780 & 207.1100 & 212.5800 & 19.6870 \\\hline
26 & 213.0780 & 208.3101 & 212.5800 & 80.6570 \\\hline
27 & 213.0780 & 208.1756 & 212.5800 & 72.3360 \\\hline
28 & 213.0780 & 208.7448 & 212.5800 & 86.4840 \\\hline
29 & 213.0780 & 207.4372 & 212.5800 & 68.5070 \\\hline
30 & 213.0780 & 207.6586 & 212.5810 & 53.3570 \\\hline
31 & 213.0780 & 208.9853 & 213.0780 & 47.4280 \\\hline
32 & 213.0780 & 209.1676 & 213.0780 & 87.2780 \\\hline
33 & 213.0780 & 209.3081 & 213.0780 & 65.1870 \\\hline
34 & 213.0780 & 208.5716 & 213.0780 & 81.0800 \\\hline
35 & 213.0780 & 209.4416 & 213.0780 & 86.9420 \\\hline
36 & 213.0780 & 209.6108 & 213.0780 & 94.2090 \\\hline
37 & 213.0780 & 207.9522 & 213.0780 & 88.4660 \\\hline
38 & 213.0780 & 208.8673 & 213.0780 & 82.1950 \\\hline
39 & 213.0780 & 209.2892 & 213.0780 & 78.5060 \\\hline
40 & 213.0780 & 209.1963 & 213.0780 & 90.2960 \\\hline
41 & 213.0780 & 208.9443 & 213.0780 & 53.6670 \\\hline
42 & 213.0780 & 209.9130 & 213.0780 & 97.0030 \\\hline
43 & 213.0780 & 210.1720 & 213.0780 & 80.7370 \\\hline
44 & 213.0780 & 209.1341 & 213.0780 & 66.9590 \\\hline
45 & 213.0780 & 209.7414 & 213.0780 & 94.2290 \\\hline
46 & 213.0780 & 209.5775 & 213.0780 & 75.3800 \\\hline
47 & 213.0780 & 209.2881 & 213.0780 & 62.2510 \\\hline
48 & 213.0780 & 207.3394 & 213.0780 & 55.4070 \\\hline
49 & 213.0780 & 208.7003 & 213.0780 & 64.2560 \\\hline
50 & 213.0780 & 208.5395 & 213.0780 & 67.6080 \\\hline
51 & 213.0780 & 210.0775 & 213.0780 & 76.2040 \\\hline
52 & 213.0780 & 209.8513 & 213.0780 & 104.1510 \\\hline
53 & 213.0780 & 207.5894 & 213.0780 & 58.4190 \\\hline
54 & 213.0780 & 208.4035 & 213.0780 & 65.1180 \\\hline
55 & 213.0780 & 209.8872 & 213.0780 & 91.0490 \\\hline
56 & 213.0780 & 208.7398 & 213.0780 & 66.3370 \\\hline
57 & 213.0780 & 208.8153 & 213.0780 & 51.8390 \\\hline
58 & 213.0780 & 209.8934 & 213.0780 & 102.0170 \\\hline
59 & 213.0780 & 209.9831 & 213.0780 & 97.9000 \\\hline
60 & 213.0780 & 209.8495 & 213.0780 & 91.8440 \\\hline
61 & 213.0780 & 208.2603 & 213.0780 & 59.0240 \\\hline
62 & 213.0780 & 208.0600 & 213.0780 & 55.4390 \\\hline
63 & 213.0780 & 209.2241 & 213.0780 & 99.2040 \\\hline
64 & 213.0780 & 208.2635 & 213.0780 & 48.9390 \\\hline
65 & 213.0780 & 209.5460 & 213.0780 & 67.5290 \\\hline
66 & 213.0780 & 209.5230 & 213.0780 & 65.9540 \\\hline
67 & 213.0780 & 209.5824 & 213.0780 & 106.7080 \\\hline
68 & 213.0780 & 208.9976 & 213.0780 & 85.6810 \\\hline
69 & 213.0780 & 209.9972 & 213.0780 & 76.7350 \\\hline
70 & 213.0780 & 208.2465 & 213.0780 & 58.3170 \\\hline
71 & 213.0780 & 209.1047 & 213.0780 & 56.5720 \\\hline
72 & 213.0780 & 208.5702 & 213.0780 & 81.9590 \\\hline
73 & 213.0780 & 209.5749 & 213.0780 & 51.7560 \\\hline
74 & 213.0780 & 209.2325 & 213.0780 & 75.8170 \\\hline
75 & 213.0780 & 210.4516 & 213.0780 & 109.7990 \\\hline
76 & 213.0780 & 209.5910 & 213.0780 & 82.4260 \\\hline
77 & 213.0780 & 208.5517 & 213.0780 & 62.0490 \\\hline
78 & 213.0780 & 209.8357 & 213.0780 & 114.0160 \\\hline
79 & 213.0780 & 209.1665 & 213.0780 & 70.5430 \\\hline
80 & 213.0780 & 209.9528 & 213.0780 & 82.5720 \\\hline
81 & 213.0780 & 209.2032 & 213.0780 & 70.5080 \\\hline
82 & 213.0780 & 208.8660 & 213.0780 & 77.5370 \\\hline
83 & 213.0780 & 208.8930 & 213.0780 & 79.4090 \\\hline
84 & 213.0780 & 209.5356 & 213.0780 & 87.1000 \\\hline
85 & 213.0780 & 208.9170 & 213.0780 & 70.5370 \\\hline
86 & 213.0780 & 209.4620 & 213.0780 & 69.1830 \\\hline
87 & 213.0780 & 208.7253 & 213.0780 & 19.0900 \\\hline
88 & 213.0780 & 209.3577 & 213.0780 & 68.2790 \\\hline
89 & 213.0780 & 209.5794 & 213.0780 & 88.4210 \\\hline
90 & 213.0780 & 209.0149 & 213.0780 & 51.8830 \\\hline
91 & 213.0780 & 209.2939 & 213.0780 & 73.1110 \\\hline
92 & 213.0780 & 208.8266 & 213.0780 & 66.2620 \\\hline
93 & 213.0780 & 208.9724 & 213.0780 & 95.3930 \\\hline
94 & 213.0780 & 209.5951 & 213.0780 & 86.2930 \\\hline
95 & 213.0780 & 209.6566 & 213.0780 & 90.7690 \\\hline
96 & 213.0780 & 210.0423 & 213.0780 & 117.8560 \\\hline
97 & 213.0780 & 210.2498 & 213.0780 & 130.5350 \\\hline
98 & 213.0780 & 208.5634 & 213.0780 & 61.4980 \\\hline
99 & 213.0780 & 207.5033 & 213.0780 & 31.5710 \\\hline
}

\resultstable{LGP Mountain Car Aggregated Results}{tab:mountain-car-lgp}{%
0 & -147.2600 & -198.8564 & -200.0000 & -200.0000 \\\hline
1 & -140.7800 & -198.0396 & -200.0000 & -200.0000 \\\hline
2 & -135.9000 & -196.7488 & -200.0000 & -200.0000 \\\hline
3 & -136.2400 & -194.4066 & -200.0000 & -200.0000 \\\hline
4 & -130.4800 & -191.1292 & -200.0000 & -200.0000 \\\hline
5 & -122.5800 & -186.1694 & -198.1600 & -200.0000 \\\hline
6 & -113.8000 & -177.7398 & -186.8600 & -200.0000 \\\hline
7 & -116.0000 & -167.0998 & -167.3400 & -200.0000 \\\hline
8 & -114.1600 & -156.7100 & -153.9800 & -200.0000 \\\hline
9 & -111.8000 & -153.0290 & -147.7200 & -200.0000 \\\hline
10 & -111.2200 & -147.6408 & -141.7600 & -200.0000 \\\hline
11 & -110.4200 & -144.4458 & -136.4100 & -200.0000 \\\hline
12 & -110.6800 & -143.9098 & -134.3500 & -200.0000 \\\hline
13 & -111.0800 & -140.9800 & -131.3100 & -200.0000 \\\hline
14 & -111.4200 & -139.6036 & -130.2800 & -200.0000 \\\hline
15 & -109.3800 & -140.3138 & -129.0800 & -200.0000 \\\hline
16 & -109.8800 & -138.5900 & -128.2600 & -200.0000 \\\hline
17 & -110.8000 & -136.8444 & -127.5100 & -200.0000 \\\hline
18 & -107.9600 & -138.1884 & -127.3900 & -200.0000 \\\hline
19 & -108.6200 & -138.4434 & -126.6200 & -200.0000 \\\hline
20 & -110.2000 & -135.1152 & -126.1500 & -200.0000 \\\hline
21 & -109.6400 & -136.9396 & -126.4000 & -200.0000 \\\hline
22 & -108.0600 & -135.8072 & -126.3900 & -200.0000 \\\hline
23 & -107.4400 & -137.5702 & -126.5300 & -200.0000 \\\hline
24 & -108.2200 & -136.7758 & -125.8500 & -200.0000 \\\hline
25 & -107.3600 & -135.2724 & -126.0000 & -200.0000 \\\hline
26 & -107.4600 & -134.7038 & -125.5000 & -200.0000 \\\hline
27 & -107.2400 & -135.3478 & -124.1200 & -200.0000 \\\hline
28 & -107.2400 & -133.9520 & -123.5200 & -200.0000 \\\hline
29 & -107.2400 & -135.3022 & -123.9400 & -200.0000 \\\hline
30 & -108.1200 & -132.1862 & -123.8800 & -200.0000 \\\hline
31 & -107.3600 & -131.6392 & -123.2600 & -200.0000 \\\hline
32 & -107.1800 & -132.2048 & -123.6200 & -200.0000 \\\hline
33 & -107.4200 & -131.1670 & -123.2400 & -200.0000 \\\hline
34 & -107.3200 & -132.0660 & -123.1600 & -200.0000 \\\hline
35 & -107.1600 & -131.6048 & -123.0800 & -200.0000 \\\hline
36 & -107.9000 & -129.9580 & -122.4500 & -197.7200 \\\hline
37 & -106.8800 & -131.1972 & -122.2800 & -200.0000 \\\hline
38 & -108.2200 & -129.5170 & -122.4700 & -200.0000 \\\hline
39 & -106.8000 & -129.6010 & -122.0600 & -200.0000 \\\hline
40 & -107.1000 & -130.2330 & -122.2300 & -200.0000 \\\hline
41 & -108.1000 & -129.9540 & -122.2500 & -200.0000 \\\hline
42 & -106.9000 & -129.8734 & -122.3600 & -197.3800 \\\hline
43 & -106.9000 & -129.6038 & -122.0700 & -200.0000 \\\hline
44 & -107.3600 & -130.1622 & -122.0300 & -200.0000 \\\hline
45 & -106.9800 & -129.5466 & -122.4000 & -200.0000 \\\hline
46 & -107.1200 & -130.2848 & -122.1200 & -200.0000 \\\hline
47 & -106.8400 & -130.3336 & -122.3300 & -200.0000 \\\hline
48 & -106.9800 & -128.9738 & -122.2900 & -200.0000 \\\hline
49 & -107.3400 & -129.2868 & -123.1800 & -200.0000 \\\hline
50 & -107.0200 & -130.7394 & -122.9600 & -200.0000 \\\hline
51 & -107.1200 & -130.8232 & -122.9500 & -200.0000 \\\hline
52 & -107.8800 & -130.4052 & -122.2000 & -200.0000 \\\hline
53 & -107.1600 & -129.0214 & -121.7800 & -200.0000 \\\hline
54 & -107.1400 & -129.8846 & -122.2000 & -200.0000 \\\hline
55 & -106.7800 & -130.4204 & -122.9700 & -200.0000 \\\hline
56 & -106.9000 & -130.6972 & -122.4100 & -200.0000 \\\hline
57 & -107.3400 & -127.9828 & -121.3700 & -200.0000 \\\hline
58 & -107.0800 & -129.4644 & -122.0200 & -200.0000 \\\hline
59 & -106.9800 & -128.7068 & -121.6100 & -200.0000 \\\hline
60 & -107.1200 & -127.6184 & -122.0900 & -200.0000 \\\hline
61 & -107.1200 & -128.2956 & -121.3300 & -200.0000 \\\hline
62 & -107.0600 & -129.2262 & -121.6300 & -200.0000 \\\hline
63 & -107.3600 & -128.8000 & -121.5700 & -200.0000 \\\hline
64 & -107.3800 & -128.2434 & -121.5400 & -200.0000 \\\hline
65 & -107.0400 & -128.0978 & -120.8800 & -199.9400 \\\hline
66 & -107.1200 & -128.2124 & -120.8200 & -200.0000 \\\hline
67 & -107.3800 & -128.7196 & -120.7600 & -200.0000 \\\hline
68 & -106.8800 & -127.0206 & -119.7800 & -200.0000 \\\hline
69 & -107.4200 & -127.9458 & -120.9300 & -200.0000 \\\hline
70 & -107.4400 & -127.6984 & -121.3600 & -200.0000 \\\hline
71 & -107.5200 & -127.6352 & -121.3900 & -200.0000 \\\hline
72 & -107.4600 & -126.7986 & -120.5000 & -200.0000 \\\hline
73 & -107.2600 & -128.4960 & -121.7400 & -200.0000 \\\hline
74 & -107.5800 & -128.9330 & -121.3100 & -200.0000 \\\hline
75 & -107.7200 & -126.9282 & -120.7300 & -200.0000 \\\hline
76 & -107.5000 & -127.7386 & -120.8500 & -200.0000 \\\hline
77 & -107.4000 & -127.4444 & -120.4500 & -200.0000 \\\hline
78 & -107.4000 & -127.3788 & -120.3000 & -200.0000 \\\hline
79 & -107.3200 & -127.6434 & -120.4100 & -200.0000 \\\hline
80 & -107.8200 & -126.4820 & -120.4000 & -199.7800 \\\hline
81 & -107.7800 & -126.8866 & -120.1500 & -200.0000 \\\hline
82 & -107.5200 & -126.9112 & -120.1400 & -200.0000 \\\hline
83 & -107.6200 & -127.6986 & -120.1900 & -200.0000 \\\hline
84 & -107.1600 & -127.9976 & -120.8700 & -200.0000 \\\hline
85 & -106.9000 & -127.3250 & -120.5200 & -200.0000 \\\hline
86 & -106.6400 & -127.1802 & -119.9500 & -200.0000 \\\hline
87 & -107.0000 & -126.6212 & -120.7200 & -200.0000 \\\hline
88 & -106.7000 & -127.0610 & -120.5600 & -200.0000 \\\hline
89 & -106.9000 & -126.3396 & -119.8800 & -196.9000 \\\hline
90 & -106.9400 & -127.0148 & -119.9500 & -200.0000 \\\hline
91 & -107.1400 & -125.9496 & -120.1000 & -200.0000 \\\hline
92 & -106.6800 & -127.0402 & -120.2600 & -200.0000 \\\hline
93 & -106.6800 & -127.4314 & -119.9800 & -200.0000 \\\hline
94 & -106.8000 & -127.3154 & -120.5300 & -198.2400 \\\hline
95 & -107.1600 & -127.7486 & -120.4700 & -200.0000 \\\hline
96 & -106.6600 & -126.7210 & -119.7800 & -200.0000 \\\hline
97 & -106.4000 & -126.7944 & -120.1900 & -200.0000 \\\hline
98 & -106.4200 & -126.5638 & -120.1500 & -200.0000 \\\hline
99 & -106.4200 & -126.8950 & -120.1000 & -200.0000 \\\hline
}

\resultstable{RLGP Mountain Car Aggregated Results}{tab:mountain-car-q}{%
0 & -159.3520 & -199.2556 & -200.0000 & -200.0000 \\\hline
1 & -148.2350 & -198.4004 & -200.0000 & -200.0000 \\\hline
2 & -129.8430 & -197.0150 & -200.0000 & -200.0000 \\\hline
3 & -125.7290 & -195.3591 & -200.0000 & -200.0000 \\\hline
4 & -124.0570 & -192.3421 & -200.0000 & -200.0000 \\\hline
5 & -121.1780 & -187.7516 & -199.4060 & -200.0000 \\\hline
6 & -120.8060 & -178.3610 & -189.7845 & -200.0000 \\\hline
7 & -120.3750 & -166.7054 & -169.4035 & -200.0000 \\\hline
8 & -119.8660 & -156.1507 & -152.4970 & -200.0000 \\\hline
9 & -119.3150 & -149.9850 & -141.1145 & -200.0000 \\\hline
10 & -119.1670 & -144.5840 & -136.2730 & -200.0000 \\\hline
11 & -119.2320 & -143.9219 & -132.7005 & -200.0000 \\\hline
12 & -118.5000 & -142.1678 & -129.3525 & -200.0000 \\\hline
13 & -118.1800 & -138.9305 & -126.1845 & -200.0000 \\\hline
14 & -118.2710 & -138.1095 & -125.1595 & -200.0000 \\\hline
15 & -117.8970 & -136.6103 & -123.5150 & -200.0000 \\\hline
16 & -117.1960 & -135.6228 & -122.9925 & -200.0000 \\\hline
17 & -117.2360 & -134.5921 & -122.7375 & -200.0000 \\\hline
18 & -117.2640 & -134.1918 & -122.2970 & -200.0000 \\\hline
19 & -117.4570 & -134.4230 & -122.2370 & -200.0000 \\\hline
20 & -117.4990 & -133.8670 & -122.0585 & -200.0000 \\\hline
21 & -117.0630 & -134.5028 & -122.0500 & -200.0000 \\\hline
22 & -117.0270 & -133.1338 & -121.7130 & -200.0000 \\\hline
23 & -116.9780 & -133.3599 & -121.7385 & -200.0000 \\\hline
24 & -116.7110 & -132.0771 & -121.4590 & -200.0000 \\\hline
25 & -117.0350 & -133.2521 & -121.2165 & -200.0000 \\\hline
26 & -116.6840 & -132.2046 & -121.0240 & -200.0000 \\\hline
27 & -116.5200 & -131.7314 & -120.9545 & -200.0000 \\\hline
28 & -116.4580 & -132.9466 & -120.8915 & -200.0000 \\\hline
29 & -116.0690 & -132.7108 & -120.6810 & -200.0000 \\\hline
30 & -116.2650 & -132.5323 & -120.5795 & -200.0000 \\\hline
31 & -116.3540 & -131.8354 & -120.5125 & -200.0000 \\\hline
32 & -115.9790 & -132.2702 & -120.6270 & -200.0000 \\\hline
33 & -116.2630 & -131.7327 & -120.4730 & -200.0000 \\\hline
34 & -116.2460 & -130.4898 & -120.4030 & -200.0000 \\\hline
35 & -116.0300 & -131.4998 & -120.3165 & -200.0000 \\\hline
36 & -115.6990 & -131.6269 & -120.3180 & -200.0000 \\\hline
37 & -116.1590 & -130.2035 & -120.2745 & -200.0000 \\\hline
38 & -115.6610 & -132.1114 & -120.4380 & -200.0000 \\\hline
39 & -116.3130 & -132.0942 & -120.2970 & -200.0000 \\\hline
40 & -116.2540 & -130.5519 & -120.1415 & -200.0000 \\\hline
41 & -116.2360 & -130.2979 & -120.1640 & -200.0000 \\\hline
42 & -116.1760 & -132.2521 & -120.1145 & -200.0000 \\\hline
43 & -116.3040 & -132.2787 & -120.2020 & -200.0000 \\\hline
44 & -116.2140 & -131.2063 & -120.1950 & -200.0000 \\\hline
45 & -115.7170 & -132.8159 & -120.0680 & -200.0000 \\\hline
46 & -116.0380 & -132.9527 & -120.1615 & -200.0000 \\\hline
47 & -116.3760 & -131.5387 & -120.0575 & -200.0000 \\\hline
48 & -115.8780 & -130.3797 & -120.0140 & -200.0000 \\\hline
49 & -116.6060 & -132.6714 & -120.1285 & -200.0000 \\\hline
50 & -115.9770 & -132.0721 & -119.8750 & -200.0000 \\\hline
51 & -115.9570 & -131.9084 & -120.0350 & -200.0000 \\\hline
52 & -115.5210 & -130.9580 & -119.8285 & -200.0000 \\\hline
53 & -115.5090 & -131.5952 & -119.9090 & -200.0000 \\\hline
54 & -115.7010 & -131.5353 & -119.6005 & -200.0000 \\\hline
55 & -115.0510 & -131.2645 & -119.6280 & -200.0000 \\\hline
56 & -115.2070 & -131.3729 & -119.4525 & -200.0000 \\\hline
57 & -115.1060 & -129.7635 & -119.4490 & -200.0000 \\\hline
58 & -114.9600 & -131.3091 & -119.5240 & -200.0000 \\\hline
59 & -115.0400 & -132.6765 & -119.4085 & -200.0000 \\\hline
60 & -114.9680 & -130.8734 & -119.1265 & -200.0000 \\\hline
61 & -114.8730 & -132.3701 & -119.2700 & -200.0000 \\\hline
62 & -114.8410 & -132.2661 & -119.1605 & -200.0000 \\\hline
63 & -115.0930 & -132.2378 & -119.2345 & -200.0000 \\\hline
64 & -115.0910 & -131.0686 & -119.0595 & -200.0000 \\\hline
65 & -115.0480 & -131.6609 & -118.9885 & -200.0000 \\\hline
66 & -114.9010 & -131.4209 & -118.9360 & -200.0000 \\\hline
67 & -114.7280 & -132.2548 & -119.0355 & -200.0000 \\\hline
68 & -114.9430 & -131.7701 & -118.8670 & -200.0000 \\\hline
69 & -114.9050 & -130.9010 & -118.8090 & -200.0000 \\\hline
70 & -114.7890 & -131.2397 & -118.8795 & -200.0000 \\\hline
71 & -114.9100 & -132.4739 & -118.7410 & -200.0000 \\\hline
72 & -114.6250 & -130.5194 & -118.7545 & -200.0000 \\\hline
73 & -114.6630 & -132.9116 & -118.7225 & -200.0000 \\\hline
74 & -114.5000 & -131.3717 & -118.1790 & -200.0000 \\\hline
75 & -114.4460 & -131.3647 & -118.0155 & -200.0000 \\\hline
76 & -114.8400 & -132.0535 & -118.0045 & -200.0000 \\\hline
77 & -114.5830 & -132.8627 & -117.9610 & -200.0000 \\\hline
78 & -114.7780 & -132.8256 & -118.0235 & -200.0000 \\\hline
79 & -114.2620 & -132.6821 & -117.9460 & -200.0000 \\\hline
80 & -114.7990 & -131.5703 & -117.8335 & -200.0000 \\\hline
81 & -114.7540 & -131.3465 & -117.7980 & -200.0000 \\\hline
82 & -114.6150 & -131.1181 & -117.9350 & -200.0000 \\\hline
83 & -114.5720 & -132.0720 & -117.9550 & -200.0000 \\\hline
84 & -114.5820 & -132.5930 & -117.8270 & -200.0000 \\\hline
85 & -114.2910 & -132.3195 & -117.7830 & -200.0000 \\\hline
86 & -114.7610 & -130.8199 & -117.8655 & -200.0000 \\\hline
87 & -114.3940 & -131.5791 & -117.8385 & -200.0000 \\\hline
88 & -114.3350 & -132.4481 & -117.8670 & -200.0000 \\\hline
89 & -114.6280 & -131.8264 & -117.6790 & -200.0000 \\\hline
90 & -114.5470 & -132.0439 & -117.6940 & -200.0000 \\\hline
91 & -114.6210 & -130.1758 & -117.7770 & -200.0000 \\\hline
92 & -114.4550 & -129.9456 & -117.6665 & -200.0000 \\\hline
93 & -114.6360 & -132.3662 & -117.7400 & -200.0000 \\\hline
94 & -114.6200 & -130.9562 & -117.6490 & -200.0000 \\\hline
95 & -114.5020 & -130.7770 & -117.6440 & -200.0000 \\\hline
96 & -114.5200 & -130.8951 & -117.6940 & -200.0000 \\\hline
97 & -114.3920 & -131.5668 & -117.7350 & -200.0000 \\\hline
98 & -114.2070 & -131.2721 & -117.6045 & -200.0000 \\\hline
99 & -114.4430 & -130.2522 & -117.6685 & -200.0000 \\\hline
}

\chapter{Conclusion}

Due to the amount of time spent on reconfiguring the framework in addition to the short nature of this research project,
various forms of experimentation did not occur. With the further optimization of hyperparameters and a different experimental setup, its possible that results could have improved. Nonetheless, the framework now exists for experimenting with easy, and opens a door for further exploration into the subject at hand. In the future, we would like to
explore different benchmarking approaches and different types of tasks besides those found in the classical control module \cite{1606.01540}, as it might bring forth insight about the types of problems where RLGP can thrive.

\bibliographystyle{plain}
\nocite{*}
\bibliography{thesis}

\end{document}